\documentclass{article}
\usepackage{ijcai21}

\usepackage{times}

\usepackage{soul}
\usepackage{url}
\usepackage[hidelinks]{hyperref}
\usepackage[utf8]{inputenc}
\usepackage[small]{caption}
\usepackage{graphicx}
\usepackage{amsmath}
\usepackage{booktabs}
\usepackage{multirow}
\usepackage{amsthm}
\usepackage{algorithm}
\usepackage{algorithmic}
\usepackage{verbatim}
\usepackage{amsfonts,amssymb,mathrsfs}
\usepackage[subfigure]{tocloft}
\usepackage{subfigure}
\usepackage{xcolor}
\newcommand{\rom}[1]{%
  \textup{\uppercase\expandafter{\romannumeral#1}}%
}
\newcommand{\jyli}[1]{\textcolor{blue}{[jyli: {#1}]}}

\title{Learning from Multiple Annotators by Incorporating Instance Features}

\author{
Jingzheng Li$^1$
\and
Hailong Sun$^1$\and
Jiyi Li$^{2}$\and
Zhijun Chen$^1$\and
Renshuai Tao$^1$\And
Yufei Ge$^3$
\affiliations
$^1$Beihang University\\
$^2$University of Yamanashi\\
$^3$Northeast Normal University\\
\emails
lijz19@act.buaa.edu.cn,
sunhl@buaa.edu.cn,
jyli@yamanashi.ac.jp,
\{zhijunchen, rstao\}@buaa.edu.cn,
geyf548@nenu.edu.cn
}

\begin{document}

\maketitle

\begin{abstract}
Learning from multiple annotators aims to induce a high-quality classifier from training instances, where each of them is associated with a set of possibly noisy labels provided by multiple annotators under the influence of their varying abilities and own biases.
In modeling the probability transition process from latent true labels to observed labels, most existing methods adopt class-level confusion matrices of annotators that observed labels do not depend on the instance features, just determined by the true labels.
It may limit the performance that the classifier can achieve. 
In this work, we propose the noise transition matrix, which incorporates the influence of instance features on annotators' performance based on confusion matrices. 
Furthermore, we propose a simple yet effective learning framework, which consists of a classifier module and a noise transition matrix module in a unified neural network architecture. 
Experimental results on one real datasets demonstrate the superiority of our method in comparison with state-of-the-art methods.
\end{abstract}

\section{Introduction}
The success of supervised learning applications often relies on large-scale well-labeled datasets.
Unfortunately, obtaining high-quality annotations from experts can be costly in terms of time and money.
Alternatively, crowdsourcing provides an inexpensive approach to data labeling by hiring world-wide annotators on public platforms like Amazon Mechanical Turk (AMT).
However, crowdsourced labels are usually noisy due to the existence of inexperienced or malicious annotators.
Using these noisy labels in supervised learning may result in an inaccurate classifier.
A straightforward way to solve this problem is redundant labeling, {\it i.e.,} obtaining multiple labels for each instance from multiple annotators.
Hence this raises one fundamental
problem termed as Learning from Crowds (LFC)~\cite{Rodrigues}: {\it ``How can we learn a good classifier from a set of possibly noisy labeled data provided by multiple annotators?''}
To address the above issue, a two-stage approach is commonly adopted. First, in {\it answer aggregation} stage  ~\cite{liguoliang,zhangjing,jin2020technical}, the latent true labels are estimated. Then, a classifier is trained based on the estimated true labels.
Alternatively, the one-stage approach~\cite{RAYKAR,tanno2019learning} 
has been shown to be a promising direction that presents a maximum-likelihood estimator that jointly learns the classifier, abilities of multiple annotators, and the latent true labels.
\begin{figure}    
\centering
\includegraphics[width=0.85\linewidth,scale=1.00]{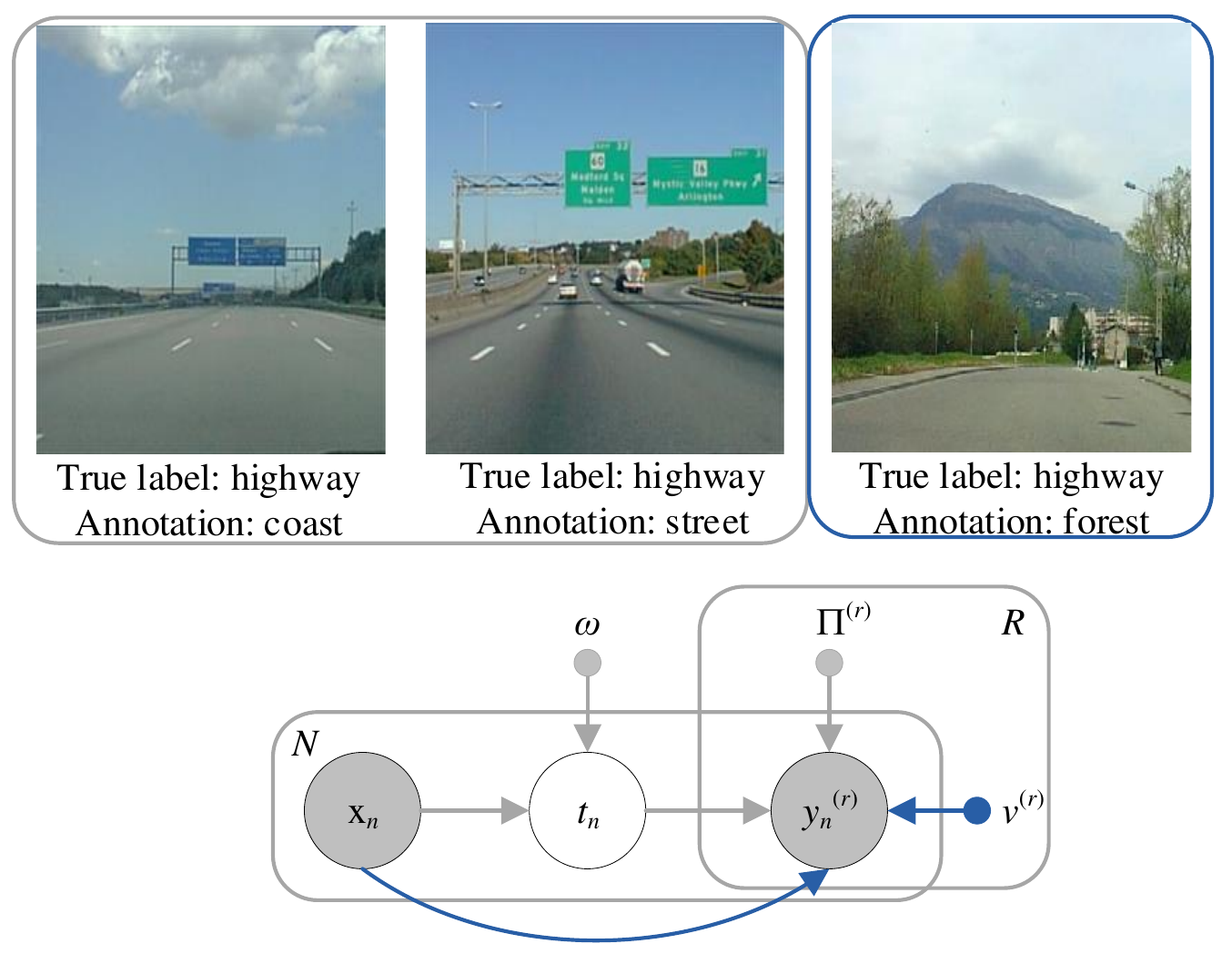}
\caption{Top: An example describes various incorrect annotations. The first randomly flips the true label to one of other classes; the second is that the true label is corrupted to the relevant class according to a fixed probability; the third is that the true label is corrupted to the irrelevant class due to the influence of instance features. Bottom: The factor graph of LFC-$\mathbf{x}$ represents the correlation of the instance $\mathbf{x}_n$, the true label $t_n$, and crowdsourced labels $y_n$. The annotation depends not only on the true label but also on instance features. }
\label{Figure:1}
\end{figure}
Among various research efforts on LFC, the probability transition process from latent true labels to observed crowdsourced labels is usually modeled with {\it confusion matrices} of annotators, which represents class-level probability transition. This means that the annotator's performance is consistent across different instances within the same class, {\it i.e.,} the transition from class $j$ to class $l$ is independent of instance features.
However, in the real world, the difficulty of labeling can vary among instances within the same class and the instance features themselves will affect annotators' performance.
Consider a concrete instantiation of LabelMe dataset~\cite{Rodrigues3} in Figure~\ref{Figure:1}.(Top), which illustrates various cases of incorrect annotations given the true label ``highway''.
The first indicates an inexperienced/malicious annotator who gives a random label ``coast''; the second indicates
an annotator have biased understanding on different classes, preferring to label ``highway'' as ``street'', because there is a strong correlation between those two classes.
In both cases, the class-level confusion matrix of annotator can be used to characterize their varying abilities and own biases.
Nevertheless, the third depicts one instance in class ``highway'' contains related visual features of other classes, misleading annotators label it as ``forest'', although these two classes are irrelevant.
We argue that the class-level confusion matrices cannot completely characterize the performance of multiple annotators across different instances within the same class.
This would limit the ability to estimate latent true labels, resulting in sub-optimal performance of the classifier.
It is necessary to consider the influence of instance features in the process of characterizing performance of multiple annotators for LFC.

To address the aforementioned deficiency, this work aims at proposing a novel LFC framework, LFC-$\mathbf{x}$, which can learn a classifier directly from crowdsourced labels provided by multiple annotators. In particular, beyond confusion matrices, LFC-$\mathbf{x}$ models the probability transition process with {\it noise transition matrices} by combining the confusion matrices and instance features.
To this end, we need to deal with two practical challenges.
One is how to quantify the influence of instance features on the performance of annotators in order to construct the noise transition matrix, the other is how to incorporate the noise transition matrix into LFC method.
To cope with these challenges, 
first, we model the correlation among instance features, latent true labels and crowdsourced labels in the probabilistic graphical model to construct the noise transition matrix. Furthermore, the LFC-$\mathbf{x}$ consists of two modules: the noise transition matrix module and the classifier module.
These two modules are integrated into an end-to-end neural network system through a principled combination for maximizing a likelihood function.
The graphical model of the LFC-$\mathbf{x}$ is presented in Figure~\ref{Figure:1}.(bottom).

\section{Related Work}
There are mainly two lines of efforts on learning
a classifier from crowdsourced labels provided by multiple
annotators.
\newline\textbf{Answer Aggregation:}
Two-stage approaches first infer true labels with {\it Answer aggregation}~\cite{liguoliang,zhangjing}, then learn a classifier.
One of the pioneer works is the DS model~\cite{Dawid}, which applies the EM algorithm to estimate latent true labels and confusion matrices of annotators.
On this basis,
Whitehill et al.~\shortcite{Jacob} consider the generalized DS model, which involves the difficulty of each instance without using instance features.
By analogy to ensemble learning, Kim and Ghahramani~\shortcite{kim2012bayesian} and Li et al.~\shortcite{LIYUAN} formalize the answer aggregation problem as Bayesian classifier combination that is capable of capturing correlations between different annotators.
Nevertheless, two-stage approaches do not realize the full potential of combining answer aggregation and classifier~\cite{khetan2017learning}.
\newline\textbf{One-stage Approaches:}
Raykar et al.~\shortcite{RAYKAR} come up with the one-stage approach, which implements an EM algorithm to jointly model abilities of annotators and learn a logistic regression classifier.
This line of work is further extended to other types of models such as convolutional neural networks~\cite{Albarqouni} and supervised latent Dirichlet allocation~\cite{Rodrigues3}. 
Of particular interest, \citeauthor{kajino2013clustering}~\shortcite{kajino2013clustering} notice that annotators form clusters according to their abilities, and apply clusters of annotators to resolve the LFC problem. Closer to our work, Rodrigues and Pereira et al.~\shortcite{Rodrigues} propose the Crowd Layer to train deep neural networks end-to-end directly from the noisy crowdsourced labels, using only back-propagation.
On this basis, Chen et al.~\shortcite{chenstructured} present a structured probabilistic model which incorporates the constraints of probability axioms into parameters of the
Crowd Layer.
More recently, Cao et al.~\shortcite{cao2019max} and Li et al.~\shortcite{li2020coupled} simultaneously aggregate the crowdsourced labels and learn an accurate classifier via a multi-view learning.
Chu et al.~\shortcite{chu2020learning} decompose the confusion matrix into two parts: one is commonly shared confusion matrix, and
the other one is individual confusion matrix.
Unlike our
method, those works are based on a common assumption: the crowdsourced labels and the instance features are independent
conditioning on the true labels.

To our knowledge, some relevant works also focus on the influence of instance features in modeling probability transition process from latent true labels to observed labels~\cite{sukhbaatar2014training,Goldberger1,chen2020beyond,zhang2021learning}.
There are two main differences between our work and these efforts.
First, these methods consider annotation scenarios where each instance corresponding to one annotation without worker information and missing annotations.
Whereas our method needs to consider the aggregation process of multiple annotations on each instance.
Second, the fine-grained instance-level noise transition matrix we construct is based on the class-level confusion matrix and it does not ignore prior class-level probability transition information thus yielding more stable and superior result.

\section{Method}
\subsection{Notation and Problem Formulation}
\label{sec:3.1}
Given $N$ instances $\{ \mathbf{x}_{1},..., \mathbf{x}_{N}\}$, in which $\mathbf{x}_{n} \in \mathbb{R}^{(D)}$ (the $D$-dimensional feature vector), that are independently sampled. Let $y_{n}^{(r)}$ represents the annotation/label for $\mathbf{x}_{n}$ provided by annotator $r$ in a set of $R$ annotators.
Formally, we set the matrix $\mathbf{X} = [\mathbf{x}_{1}^T;...;\mathbf{x}_{N}^T] \in  \mathbb{R}^{(N \times D)}$ and $\mathbf{Y} = [y_{1}^{(1)},...,y_{1}^{(R)};...;y_{N}^{(1)},...,y_{N}^{(R)}] \in \mathbb{R}^{(N \times R)}$ in which $(\cdot)^T$ represents matrix transposition. 
We denote the latent true labels for $\mathbf{X}$ by $ \mathbf{T} = [t_{1};...;t_{N}]$.
Given the observed training data $\mathbf{X}$ and $\mathbf{Y}$, 
the goal of interest is to jointly estimate abilities of multiple annotators and latent true labels, along with an accurate classifier.

In general learning, there are two common assumptions: 1) multiple annotators provide crowdsourced labels that are independent given input instances; 2) crowdsourced labels do not depend on the instance features, just determined by the true labels. 
Conditioning on the true labels, 
the probability of crowdsourced labels on instance features can be factored as
\begin{equation}
     p(\mathbf{Y} | \mathbf{X}, \Theta )=\prod_{n=1}^{N} \sum_{t_{n}} p(t_{n} | \mathbf{x}_{n}, \boldsymbol{w}) \prod_{r=1}^{R} p(y_{n}^{(r)} | t_{n}, \boldsymbol{\Pi}^{(r)}).
\label{eq:1}
\end{equation}
where $p(t_{n} | \mathbf{x}_{n}, \boldsymbol{w})$ represents true label distribution parameterized by $\boldsymbol{w}$, and $p(y_{n}^{(r)} | t_{n}, \boldsymbol{\Pi}^{(r)})$ parameterized by matrix $\boldsymbol{\Pi}$ depicts the class-level probability transition that the annotator, $r$, will annotate class $y_{n}^{(r)}$ given true label $t_{n}$. Specifically,
the matrix $\mathbf{\Pi}^{(r)} = (\pi_{i j}^{(r)})_{C \times C} \in[0,1]^{(C \times C)}$ is called the {\it confusion matrix} for representing the $r$-th annotator's ability whose $(i, j)^{\mathrm{th}}$ element is parameterized by $\pi_{i j}^{(r)}$ where $i, j \in \{1, \ldots, C\}$ and $C$ is the number of classes.
To achieve the goal, the EM algorithm or neural network ~\cite{Rodrigues} can be used to compute the maximum-likelihood solution, formalized as
\begin{equation}
    \hat{\Theta}_{\mathrm{ML}}=\arg \max _{\Theta}p(\mathbf{Y} | \mathbf{X}, \Theta).
\label{eq:2}
\end{equation}

According to the previous empirical analysis of the probability transition process from true labels to crowdsourced labels, crowdsourced labels depend not only on true labels but also on instance features. 
Modeling the probability transition process considering only confusion matrices of multiple annotators would limit the ability to infer the latent true labels and lead to sub-optimal performance of the classifier.
Consequently, it is necessary to consider the influence of instance features to characterize the probability transition process.

\subsection{Proposed  LFC-$\mathbf{x}$}
In our setting, we here make a key assumption that the crowdsourced labels depend on not only true labels but also instance features.
We propose the {\it noise transition matrix} to model probability transition process on multiple annotators given input instances through incorporating input instance features into confusion matrix.
Now we rewrite the probability of given crowdsourced labels, then
we take its log and the part of the probability transition in Equation~\ref{eq:1} related to confusion matrix is
replaced with the noise transition matrix, yielding
\begin{equation}
\resizebox{.9\linewidth}{!}{$
    \displaystyle
\begin{aligned}
    &\log p(\mathbf{Y} | \mathbf{X}, \Theta ) = \sum_{n=1}^{N} \sum_{r=1}^{R} \log p(\mathbf{Y}^{(r)} | \mathbf{X}, \Theta^{(r)} ) \\
    &= \sum_{n=1}^{N} \sum_{r=1}^{R} \log \sum_{t_{n}} \underbrace{p(t_{n} | \mathbf{x}_{n}, \boldsymbol{w})}_{\text {classifier}} \underbrace{p(y_{n}^{(r)} | t_{n},\mathbf{x}_{n}, \boldsymbol{\Pi}^{(r)}, \boldsymbol{v}^{(r)})}_{\text {noise transition matrix}},
\end{aligned}
$}
\label{eq:3}
\end{equation}
where the $\Theta$ is a collection of $\{\boldsymbol{w}, \{\boldsymbol{\Pi}^{(r)}\}_{r=1}^{R}, \{\boldsymbol{v}^{(r)}\}_{r=1}^{R}\}\}$, and the parameter $\{\boldsymbol{v}^{(r)}\}_{r=1}^{R}$ represents the influence of instance features themselves on workers’ performance that we will discuss later.

To instantiate our probabilistic graphical model as shown in Figure~\ref{Figure:1}.(bottom), we propose LFC-$\mathbf{x}$ that minimizes the negative log-likelihood function with respect to the parameter $\Theta$ within the framework of neural networks that consists of two modules: a {\it classifier module} and a {\it noise transition matrix module}. 
Specifically, the log-likelihood is the output of LFC-$\mathbf{x}$ through a principled combination of the classifier module and the noise transition matrix module.
Next, we describe how to jointly optimize parameters of the classifier module and the noise transition matrix module.
\newline\textbf{Classifier Module:}
Without loss of generality, suppose that a  softmax neural classifier is given,
parameterized by $\boldsymbol{w}$, for inferring the true label distribution.
We denote the non-linear function applied to an instance $\mathbf{x}_n$ by $h(\mathbf{x}_n)$ for extracting the instance features, and denote the softmax layer that predicts the true label $t_n$ by
\begin{equation}
    p(t_n=j | \mathbf{x}_n, \boldsymbol{w})=\frac{\exp (\mathbf{u}_{j}^{\top} h(\mathbf{x}_n)+b_{j})}{\sum_{i=1}^{C} \exp (\mathbf{u}_{i}^{\top} h(\mathbf{x}_n)+b_{i})},
\label{eq:4}
\end{equation}
in which the vector $\mathbf{u}$ and scalar $b$ indicate weight and bias, respectively, and $i, j \in \{1, \ldots, C\}$.
Nevertheless, we do not have access to latent true labels so that the classifier module alone is not enough.
Therefore, a noise transition matrix module that can model the labeling process of multiple annotators, {\it i.e.,} the  probability transition from true labels and instance features to crowdsourced labels, is required to provide weakly supervised information for classifier.
In doing so, the classifier can be trained by multiplying its output by the estimated noise transition matrices which is shown in Equation~\ref{eq:3}.
\newline\textbf{Noise Transition Matrix Module:} 
Differ from worker-specific confusion matrix that characterizes class-level probability transition, we propose worker-specific noise transition matrix that characterizes instance-level probability transition through incorporating instance features into confusion matrix.
Now, a major challenge is how to construct the noise transition matrix.
Naturally, it contains two problems. The first one is how to quantify the impact of instance features on annotators' performance.
The second one is how to incorporate such 
impact of instance features into confusion matrix for constructing noise transition matrix.
%

The first problem comes from how to specify the mapping function from instance features to the annotators' performance.
To this end, we construct {\it instance impact matrix} parameterized by $\boldsymbol{v}^{(r)}$ to characterize the instance features' impact on annotators' performance.
More concretely, we explore a solution by adding a linear layer with $C^2$ units on top of instance features in the classifier module for each worker.
Recall that the instance features are extracted by the non-linear function $h(\mathbf{x}_n)$ defined in Equation~\ref{eq:4}, which is the output of penultimate layer of classifier.
This linear layer without bias called {\it instance impact matrix layer} is parallel with the softmax output layer of the classifier module, formalized as
\begin{equation}
    f(\mathbf{x}_n)^{(r)} = \boldsymbol{v}^{(r)} h(\mathbf{x}_n),
\label{eq:5}
\end{equation}
where the vector $\boldsymbol{v}^{(r)}$ is the weight of instance impact matrix layer.

Now that we obtain the instance impact matrix, the other crucial point is to construct the noise transition matrix.
To solve this issue, we make an intuitive assumption that the noise transition matrix is the result of addition of the instance impact matrix on the confusion matrix.
To be computationally feasible, we convert instance impact matrix layer from vector to matrix with the same shape as confusion matrix, then add them up, followed by a softmax operation, yielding
\begin{equation}
\resizebox{.98\linewidth}{!}{$
    \displaystyle
    p(y_n^{(r)}=j | t_n=i, \mathbf{x}_n, \boldsymbol{\Pi}^{(r)}, \boldsymbol{v}^{(r)}) = \frac{\exp (f(\mathbf{x}_n)_{ij}^{(r)} + \pi_{ij}^{(r)})}{\sum_{k=1}^{C} \exp ( f(\mathbf{x}_n)_{ik}^{(r)} + \pi_{ik}^{(r)})}.
$}
\label{eq:6}
\end{equation}

With all modules ready, the LFC-$\mathbf{x}$ integrates a classifier module and a noise transition matrix module into a unified neural network in an end-to-end manner for jointly estimating true labels, abilities of annotators, and learning a classifier.

We take a closer look at the optimization objective of LFC-$\mathbf{x}$ as Equation~\ref{eq:3}. 
The penultimate layer (extracted instance features) of the classifier module is shared among every annotator and becomes an ``information hub'' that connects the noise transition matrix of each annotator.
Then given a loss function $\mathcal{L}(p(y_n^{(r)} | \mathbf{x}_{n}, \Theta^{(r)}), y_n^{(r)})$ such as Cross Entropy between the model outputs and the crowdsourced labels,
minimizing the negative log-likelihood encourages the outputs of LFC-$\mathbf{x}$ to be as close as possible to the observed crowdsourced labels.
In doing so, we can perform back-propagation end-to-end for updating parameters in classifier module and noise transition matrix module.
Besides, the problem of missing labels from some of annotators can be addressed by setting their gradients to 0.
\subsection{Training}
There are degrees of freedom in the outputs of a classifier. In other words, the outputs of classifier may not semantically correspond to the true labels even if the negative log-likelihood function is minimized~\cite{sukhbaatar2014training}. Therefore, a reasonable initialization of noise transition matrix is crucial for successful convergence of the LFC-$\mathbf{x}$ for training a high-quality classifier.
We initialize the confusion matrix so that it has relatively large diagonal elements ({\it i.e.,} $\pi_{i i}>\pi_{i j} \text { for } \forall i, j \neq i$), and small symmetric noise in off-diagonal elements, {\it i.e.,}
\begin{equation}
    \pi_{i j}^{(r)}=\log (\epsilon^{\delta(i, j)} \times (\frac{1-\epsilon}{C-1})^{(1-\delta(i, j))}),
\label{eq:10}
\end{equation}
in which $\delta\left(a, b\right)$ is the indicator function and $\epsilon$ is set it to 0.46 for all datasets in experiments.
The parameters of instance impact matrix layer are initially set to 0.

Let us finally describe the training procedure which consists of two stages.
The first stage is to update confusion matrices. 
Concretely, we freeze the instance impact matrix layer so that the noise transition matrix degenerates into the confusion matrix since the instance impact matrix is fixed to 0. 
Then, we train the LFC-$\mathbf{x}$ for updating confusion matrices.
The second stage is to update noise transition matrices. Concretely, we unfreeze the instance impact matrix layer and retrain the LFC-$\mathbf{x}$ except for the learned confusion matrices. 
Once the LFC-$\mathbf{x}$ is trained, the classifier module can be used to make predictions alone for unseen instances.

\section{Experiments}
We evaluate our proposed LFC-$\mathbf{x}$ by comparing it with representative LFC baselines on real crowdsourcing datasets,
All experiments are implemented using the Keras framework.
\newline\textbf{Real Dataset:}
LabelMe dataset~\cite{Rodrigues3} is an image classification dataset involving 8 classes including ``highway'', ``inside city'', ``tall building'',
``street'', ``forest'', ``coast'', ``mountain'' and ``open country'', and it contains 2,688 images in total. Among them, 1,000 images are annotated by AMT annotators and each is annotated by 2.547 annotators on average. The remaining 1,688 images are used for testing.
\newline\textbf{Competing Strategies:}
Two-stage approaches with neural networks include ``\textbf{NN-MV}''and ``\textbf{NN-DS}'' ~\cite{Dawid}.
One-stage LFC approach contains ``\textbf{Crowd Layer}''~\cite{Rodrigues}. We also compare the LFC-$\mathbf{x}$ to one-stage EM-based LFC methods including
 ``\textbf{Raykar}'' ~\cite{RAYKAR} and ``\textbf{AggNet}'' ~\cite{Albarqouni}.
\newline\textbf{Experimental Setup:} 
We use a pre-trained VGG-16 as the backbone network and apply one fully connected layer with 128 units.
The batch size is set to 256 and we run 400 epochs.
The model is trained using SGD with momentum of 0.9, weight decay of $10^{-4}$, and an initial learning rate of 0.01. The learning rate
is divided by 10 after epochs 100 and 200.
%
Jiang et al.,~\shortcite{jiang2020beyond} point out that early stopping is not always effective on label noise. Therefore, we report not only the optimal test score during training, but also the test score of last epoch, to see the robustness of methods.
Each experiment is repeated 10 times independently and the integration accuracies are averaged. 
\begin{table}[t]
\centering
\begin{tabular}{c|c}  
\toprule[2pt]
Methods & Accuracy\\ \hline   
NN-MV & 80.72/77.10   \\  
NN-DS&  83.25/79.97\\
Raykar&75.58/75.08 \\
Aggnet&83.95/83.34\\
Crowd Layer&84.18/81.44\\
LFC-$\mathbf{x}$ & \textbf{85.86}/\textbf{84.60}\\
\bottomrule[2pt]
\end{tabular}
\caption{The test accuracies of the best during training
(left) and the last epoch (right) are listed on real crowdsourcing dataset. The best results are in \textbf{boldface}.}
\label{table2}
\end{table}

Table~\ref{table2} summarizes the comparisons of LFC-$\mathbf{x}$ to other LFC approaches.
The LFC-$\mathbf{x}$ is superior to all the compared LFC methods, which empirically demonstrates that we provide a realistic and applicable noise transition matrix for designing an LFC framework.

\section{Conclusion}
In this paper, 
we propose to learn a classifier from crowdsourced labels provide by multiple annotators.
Specifically, we first construct the noise transition matrix by incorporating instance features into confusion matrix. 
Furthermore, we propose LFC-$\mathbf{x}$ that integrates a classifier module and a noise transition matrix module into a unified neural network in an end-to-end manner.
The experiment shows the advantages of LFC-$\mathbf{x}$, confirming the effectiveness of noise transition matrix compared to class-level confusion matrix.
Besides, an advantage of LFC-$\mathbf{x}$ is its flexibility, the LFC-$\mathbf{x}$ is orthogonal to some label-denoising techniques \cite{song2020learning}, such as sample selection, loss adjustment, etc., where one can also combine them for even better performance.


\bibliographystyle{named}
\bibliography{ijcai21}
\end{document}